\documentclass[11pt]{article}
\usepackage{cite}
\usepackage{graphicx} 
\usepackage[top=1.2in, bottom=1.1in, left=1.5in, right=1.5in]{geometry}

\def\zapbib{I

 \end{document}}
\def\zap{\end{document}}
\def\zip{\begin{document}}
\def\mfn#1{_{\hbox{\it #1}}}
\def\bbox#1#2{{\vbox{\hbox{#1}\hbox{#2}}}}
\def\myeq#1{\begin{equation} #1 \end{equation}}
\newcommand{\cent}[1]{\begin{center} #1 \end{center}}
\def\taba#1#2{\parbox[b]{.5in}{\flushright #1:} \= \quad
		\parbox[t]{3.8in}{#2}\\} 
\def\tabb#1#2{\parbox[b]{.5in}{\flushright #1:} \>\quad
		\parbox[t]{3.8in}{#2}\\}
\def\picture #1 by #2 (#3){
  \vbox to #2{
     \hrule width #1 height 0pt depth 0pt
     \vfill
     \special{picture #3}}
}
\def\doublepicture #1 by #2 (#3 and #4) {
  \vbox to #2{
     \hrule width #1 height 0pt depth 0pt
     \vfill
     \special{picture #3} \hspace{2.55in}\special{picture #4}}
}

\def\scaledpicture #1 by #2 (#3 scaled #4){{
   \dimen0=#1 \dimen1=#2
   \divede\dimen0 by 1000 \multiply\dimen0 by #4
   \divede\dimen1 by 1000 \multiply\dimen1 by #4
   \picture \dimen0 by \dimen1 (#3 scaled #4)}}
\newcommand{\figbox}[2]{
    \begin{figure}
    \center{\framebox[3in][c]{\rule[-.5in]{0in}{1in} FIGURE #1}}
    \caption{\em #2}
    \end{figure} }

\begin{document}

\title{\bf Reinforcement Control with Hierarchical
Backpropagated Adaptive Critics\footnote{
This work was supported by 
Jameson Robotics and the Lyndon Baynes Johnson Space Center.
Submitted to {\em Neural Networks}.}
} 
\author{ John W. Jameson, Jameson
Robotics,\\ 4600 Shoalwood Ave., Austin, TX 78756}

\def\myeq#1{\begin{equation} #1 \end{equation}}
\def\LL{$\cal LL\ $} 
\def\HL{$\cal HL\ $} 

\newcommand{\tline}{\\ \hline}
\newcommand{\ttline}{\\ \hline\hline}

\newcommand{\xt}{{\bf x}_t}
\newcommand{\yt}{{\bf y}_t}
\newcommand{\Yt}{{\bf Y}_t}
\newcommand{\rh}{{\hat r}}

\newcommand{\mstrut}{\rule[-6pt]{.0in}{20pt}}
\newcommand{\bstrut}{\rule[-10pt]{.0in}{20pt}}
\newcommand{\tstrut}{\rule[-4pt]{.0in}{20pt}}

\date{March 19, 1990}
\maketitle

\begin{abstract} 
Present incremental learning methods are limited
in the ability to achieve reliable credit assignment over a large number
time steps (or events). However, this situation is typical for cases where
the dynamical system to be controlled requires relatively frequent
control updates in order to maintain stability or robustness yet has some
action/consequences which must be established over relatively long periods
of time.  To address this problem, the learning capabilities of a control
architecture comprised of two Backpropagated Adaptive Critics (BAC's) in a
two-level hierarchy with continuous actions are explored. The high-level BAC
updates less frequently than the low-level BAC and controls the latter to
some degree. The response of the low-level to high-level signals can either
be determined {\em a priori\/} or it can {\em emerge} during learning. A
general approach called {\em Response Induction Learning\/} is introduced to
address the latter case. 
\end{abstract}

\section{Introduction \label{sec-10}}
There has recently been considerable interest in the application of
reinforcement learning methods to problems of adaptive (or intelligent)
control. Perhaps the most recognized early work on this subject is a 1983
paper by  Barto, Sutton, and Anderson ~\cite{barto:1983} which demonstrated
stabilization of the oft-studied cart-pole problem (see Figure 2).  In this
paper the authors introduced the ``adaptive heuristic critic,'' the function
of which is to evaluate the current state in terms of expected future
(cumulative) reinforcements. Sutton's work on {\em temporal differences}
~\cite{sutton:1988} refined these notions, and extended their application
to nonlinear feedforward neural networks. His work is the basis for the
present ``critic network,'' the goal of which is to estimate the sum of
future reinforcements ($p_t$):

\myeq{ p_t\ =\ \sum_{k=0}^\infty \gamma^k r_{t+k+1},\label{eq-10}}

\noindent  where $r_{t+k+1}$ is the reinforcement received at time $t+k+1$
and $\gamma$ is the discount factor  ($0 < \gamma < 1$). Note that the
closer the discount factor is to one the greater the ``foresight'' of the
critic, i.e., the more future reinforcements should incorporated in the
current prediction.  
 
For many applications, in order to achieve dynamic stability and control
efficacy the interval between controller updates are  likely to be small
compared to the time between some actions and their consequences,
especially for more difficult ``intelligent'' control problems.  However,
the greater the number of time steps between initial states and goal
states, the more difficult it is to achieve reliable credit  assignment for
actions which incurred early on, i.e., the more ``foresight''  required by
the critic.  A reasonable solution is to have motion primitives which
require control actions to ``steer'' them in  some way. These steering
commands would occur less frequently than the motor commands  issued by the
primitives. Watkins ~\cite{watkins:1989} presented an interesting paradigm
for hierarchical control based on ship navigation which has several
parallels to, the present work. This idea is also similar to the principle
of ``cascade control'' developed for process control. 

In this paper a simple two-level hierarchy of reinforcement learning
modules is implemented in such a way that the ``low level'' module controls
the actuators directly and is rewarded for maintaining stability while the
``high level'' module steers the low level module in order to maximize
longer term positive reinforcement. Two cases are considered: one in which
the role of the steering command is predefined in terms of the
reinforcement to the low level controller, and the much more difficult
cases where the role is not defined a priori but is developed as learning
progresses.

\section{The Backpropagated Adaptive Critic (BAC)\label{sec-20}}

\subsection{Background}
In the Barto etal. work discussed in the last section, the state space of
the plant was broken up into ``boxes'' and the control actions were binary.
While their results were impressive, the discrete nature of the method is
impractical for problems which either require high fidelity control or
which have a high dimensional state space.  Nevertheless, they achieved
stabilization of the cart-pole system within seventy trials on
average---considerably better than any of the continuous approaches
appearing in the literature (or here).

Anderson ~\cite{anderson:1988} later used critic and action modules based on
backpropagated multilayer perceptrons for stabilization of the cart-pole
problem. Although inputs to the modules were continuous, the associative
learning method for training the action network still required (stochastic)
binary actions.  Around 8,000 trials were required on average to stabilize
the system. Note that a ``trial'' starts with the cart and pole randomly
within the state space ends when the position of either exceed its bounds.

Werbos ~\cite{werbos:1974,werbos:1990} introduced an architecture called
the {\em Backpropagated Adaptive Critic (BAC)} for which a utility function
similar to a critic is used as a guide for training the action module. In
this approach continuous inputs {\em and actions\/} are allowed.  Jordan and
Jacobs ~\cite{jordan:1989} and Jameson ~\cite{jameson:1990} provided some
of the first results with this approach, and both studied the cart-pole
problem.  Results for these cases are presented in Table 1 (and shall be
discussed further).

\subsection{Architecture: Indirect BAC}
A schematic of the BAC is shown in Fig. 1, and consists of three networks:
the {\em action,  model, and critic\/} networks, indicated by the letters
``A,'' ``M,'' and ``C,'' respectively. This architecture, for reasons to be
explained shortly, is called the {\em Indirect\/} BAC. For a more detailed
description of this architecture, see ~\cite{jameson:1990,werbos:1990}.
Note that arrows crossing the left border of the BAC in Fig. 1 correspond
to signals to and from the same BAC but for the previous time step, and
similarly arrows that cross the right border refer to the next time step.
Dashed lines represent feedback errors which may pass through a network
with or without affecting its weights;  if a dashed line is visible inside
the network block the weights are not affected, otherwise they are.  

For the experiments discussed below all the networks contained a single
hidden layer with (seven) Gaussian activation functions and an output layer
with linear activation functions.

\begin{figure}[htp]
\centering
\includegraphics[width=0.9\textwidth]{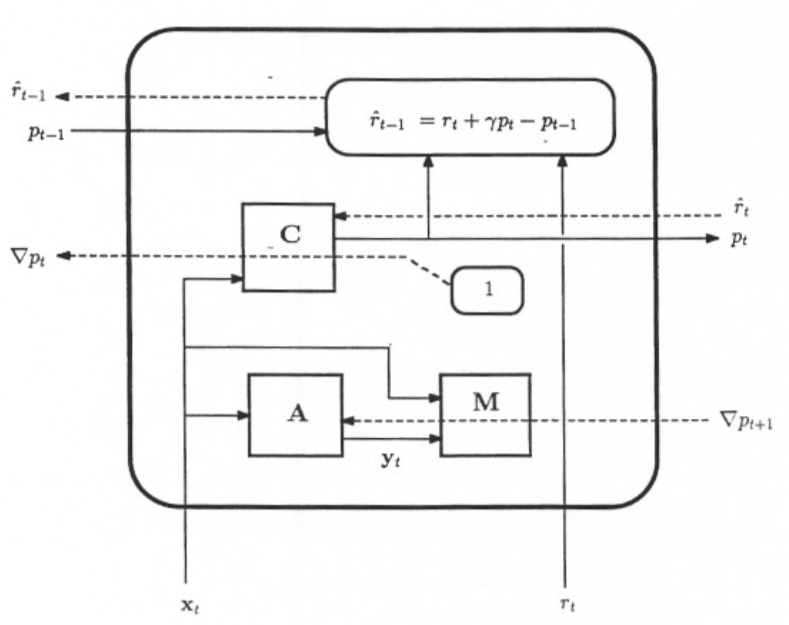}
\caption{\it BAC for one time step}\label{fig:HN}
\end{figure}

The model network performs system identification. In the preferred form, it
predicts the change in state $\Delta x$ as a function of the current state
and action, i.e., its weights are adapted to minimize, for all times,

	\myeq{k_m\ (\Delta x\ -\ \Delta x_{pred}),\label{eq-19}}

\noindent
where $\Delta x_{pred}$ is the output of the model network, and $k_m$ is a
scaling factor; the feedback path for training the model network
is not shown in Fig. 1. Training of this network is best achieved
{\em before\/} training the controller by starting the system in a random
place in the state space and letting the system evolve with random actions.
After this network has adapted satisfactory, its weights are frozen.

From eq. \ref{eq-10} it is easy to show that if the critic predictions are
correct then $p_t = r_{t+1} + \gamma p_t$.  Hence, for training the
critic,  if $r_{t+1} + \gamma p_t$ is the target and $p_t$ the actual
output, then the critic error is:

\myeq{\hat r_t \equiv r_{t+1} + \gamma p_{t+1} - p_t,\label{eq-20}}

\noindent 
which corresponds to the following equation for changing the
weights of the critic network (see ~\cite{sutton:1988} for further
clarification):

	\myeq{\Delta {\bf w}_{c,t} = \eta\ {\hat r}_{t+1} {\bf \nabla} p_t 
				\quad +\quad\hbox{momentum},\label{eq-30}}
 
\noindent 
where ${\bf w}_{c,t}$ is the weight vector for the critic
network, $\eta$ is the learning rate, and ${\bf \nabla} p_t$ is the
gradient of $p_t$ with respect to the weights.\footnote{This particular
form of the error gradient is referred to as TD(0) by Sutton
~\cite{sutton:1988}, where the ``0'' refers to the fact that no past
values of the gradient are used in eq. \ref{eq-30}.}

The action (controller) network determines which action  $\yt$ to take as
a function of the current state $\xt$ (note that in general $\yt$ is a
vector, but for the cart-pole problem it is just a scalar quantity). 
  
To adapt the action network in order to increase $p_t$,  a path is needed
from the critic network to the action network in order to feed back the
gradient of $p_t$.  However, since the critic output is a function only of
the state ($\xt$),  the model network is needed to provide a feedback
pathway to the action network.   Note that the gradient of $p_t$
corresponds to a ``error signal'' of one (1) at the output of the critic.
In order to establish more accurate gradients (and enhance
robustness) a small Gaussian noise component is added to the outputs of the
action network. However, feedback is performed with activities corresponding
to zero noise (see ~\cite{werbos:1990}).

\subsection{Direct BAC Architecture}
Another form of the BAC, called the ``Direct BAC,'' is virtually the same
as the BAC just described except that the critic input consists of the
current state and action---no model network is needed. For this case the
action is included as input to the critic network. 

\section{Single Level BAC Control with the Cart-Pole
Problem\label{sec-30}} 
\subsection{The Cart-pole System}
The cart-pole system, studied extensively in the literature in connection
with reinforcement learning, is shown in the Fig. 2.  The
state of this system is completely described by four quantities which
comprise the state vector ${\bf x}_t$: the position and velocity of the
cart, and angular position and rate of the pole. The goal is to keep the
inverted pole balanced and the cart from hitting either end of the track by
controlling the force (proportional to $y_t$) exerted on the cart. The
equations of motion of the system are:

\begin{eqnarray}
\ddot x\ &=&\ {{f_c\ +\ m_p L \left[\dot\theta^2 \sin\theta
\ -\ \ddot\theta\cos\theta\right]}\over
m_c\ +\ m_p} \label{eq-cp1}\label{eq-cp1}\\ [5pt]
\ddot\theta\ &=&\ 
{g\sin\theta\ +\ \cos\theta \left[
{-f_c\ -\ m_pL\dot\theta^2\sin\theta \over m_c + m_p}
\right]
\over 
L\left[ 
{4\over 3}
\ -\ 
{m_p\cos^2\theta\over m_c + m_p}
\right]}
\label{eq-cp2}
\end{eqnarray}

\noindent
where,
\begin{eqnarray*}
m_c\ \  =&\ \  1.0\, kg\ \  &=\ \ \hbox{mass of the cart}\\
m_p\ \  =&\ \  0.1\, kg\ \  &=\ \ \hbox{mass of the pole}\\
L\ \  =&\ \  1.0\, m\ \  &=\ \ \hbox{(nominal) length of the pole}\\
g\ \  =&\ \  9.8\, m/s^2\ \  &=\ \ \hbox{acceleration of gravity}
\end{eqnarray*}

\begin{figure}[htp]
\centering
\includegraphics[width=0.7\textwidth]{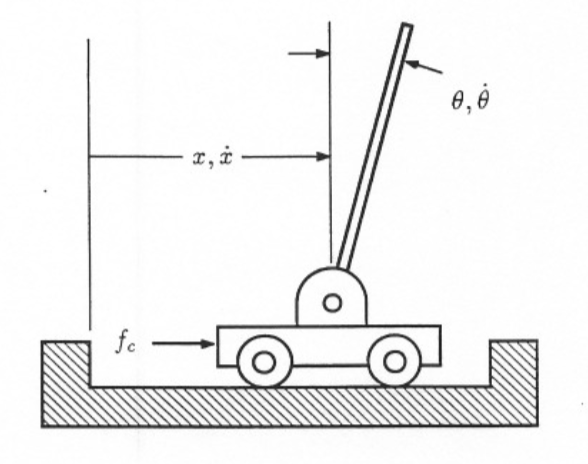}
\caption{\it The cart-pole system}\label{fig-cp}
\end{figure}

The external reinforcement signal ($r$) used for the experiments was:

\myeq{r\ =\ \sqrt{ \left({\theta\over\theta_r}\right)^2\ 
+\ \left({x\over x_r}\right)^2 },\label{eq-22}}

\noindent
where $\theta_r=12^{\small o}$ and $x_r=2.4$m are the ranges of the motion
allowed for the pole angle and cart position, respectively; if either range
is exceeded, a new trial begins. Note that I generally found the learning to
be roughly half as fast if the system was failure driven, i.e., $r=-1$ when
a failure occurs, or $r=0$ otherwise (see ~\cite{jameson:1990} for more
detailed performance comparisons). 

To balance the pole it is necessary to have this force adjusted fairly
frequently.  However, the {\em gross} motion of the cart depends on the
short-term averaged angle of the pole; if it is generally tilted to the
right the cart will slowly accelerate to the right (actually, a small
oscillatory motion which keeps the pole balanced will be superimposed on
this acceleration---note that if the pole is ``balanced'' at a constant
angle $\theta_b$, the acceleration of the cart is $\ddot x = g\tan
\theta_b$).  The point is that the time constant of this gross motion of
the cart is much longer than that of the pole. In other words, this problem
requires strong critic foresight in order to keep the cart centered.  This
fact motivated me to perform similar experiments as my earlier paper
~\cite{jameson:1990} but with a slower servo rate, i.e., all other
parameters were kept constant except for the servo rate. 

\subsection{Results for the Single-Level BAC with Different Servo Rates}
The results of these experiments  are  presented in Table 1. An experiment
consisted of letting the BAC learn until it either stabilized the system
(called a ``success'' in Table 1) or until the number of trials reached
1,200 (or 3,000 for the Direct BAC), constituting a ``failure.'' The
parameters for all the experiments the (except case 4) were identical; the
learning rates for the action and critic networks were .01 and .02,
respectively. The model network was trained in each case for 1,000
time steps.

\begin{table}
\begin{minipage}[t]{4in}
\center{
\begin{tabular}{|c|l|c|c|r|r|} \hline\hline 
\multicolumn{6}{|c|}{\tstrut {\bf TABLE 1}} \\ 
\multicolumn{6}{|c|}
			{\tstrut Experimental Results for the Cart-Pole with}\\ 
\multicolumn{6}{|c|}
			{\bstrut Single-Level Adaptive Critic Architectures} \ttline  
\mstrut Case & Architecture & servo rate & $(SR)^a$ &
$(N\mfn{ave})^b$ & 
      $(M\mfn{ave})^c$\\  
\bstrut & & (hz) & & & \tline
\mstrut 1 & Indirect BAC & 50 & 6/30 & 860 & 110,000 \tline
\mstrut 2 & Indirect BAC & 25 & 15/30 & 686 & 47,500 \tline
\mstrut 3 & Indirect BAC & 17 & 14/30 & 760 & 35,500 \ttline
\mstrut 4 & Direct BAC$^d$ & 50 & 18/20 & $\approx 20,000$ & NA \tline
\mstrut 5 & Direct BAC & 25 & 3/10 & 2,300 & 380,000 \tline
\mstrut 6 & Direct BAC & 17 & 11/30 & 1,100 & 120,000 \tline
\multicolumn{6}{l}{$^a(SR)$ = (no. of successes)/(no. of experiments)} \\
\multicolumn{6}{l}{$^b(N\mfn{ave})$ = Average no. of trials prior to
   success}\\  
\multicolumn{6}{l}{$^c(M\mfn{ave})$ = Average no. of time steps
prior to successful trial}\\ 
\multicolumn{6}{l}{$^d$from Jordan and Jacobs ~\cite{jordan:1989}}
\end{tabular}  
}
\end{minipage}
\end{table}

The results in Table 1 show that both BAC's (Indirect and Direct) had
superior performance for the faster servo rates. This was mostly likely for
the following reason: although the shortened pole is harder to balance for a
slower servo rate (the gains must be more finely tuned), less foresight is
required from the critic to center the cart. 

The data for case 4 in Table 1 was obtained from Jordan and Jacobs
~\cite{jordan:1989}, who used an architecture very similar to the Direct
BAC but with a different reinforcement signal (related to the time to
failure). However, whereas the criteria for a ``success'' was 20,000 time
steps for other cases in Table 1, only 1,000 time steps qualified as a
success in Jordan and Jacobs paper.  

Note that the control {\em logic\/} involved in stabilizing the cart-pole
system is not very complicated. The main problem seems to be the long time
constant of the cart's motion. The results of Table 1 strongly corroborate
this hypothesis.

\section{The Two-Level BAC with Explicit Low-level Role\label{sec-33}}

Put most simply, the principle of the two-level BAC is this: use a
low-level (\LL) BAC with a fast servo rate to directly control the actuators
for short term stability, and a high-level (\HL) BAC with a slower servo
rate to control the \LL BAC to maximize the (long term) reinforcement.

The preferred form of the two-level BAC is shown in Fig. 3. Only one \HL
BAC and one \LL BAC are used in this architecture.  The \HL BAC updates
once for every N \LL BAC updates, where  N was 40 for the simulations
below.\footnote{Actually, any value N between 10 and 50 works for the most
of the examples below.}

\begin{figure}[htp]
\centering
\includegraphics[width=0.65\textwidth]{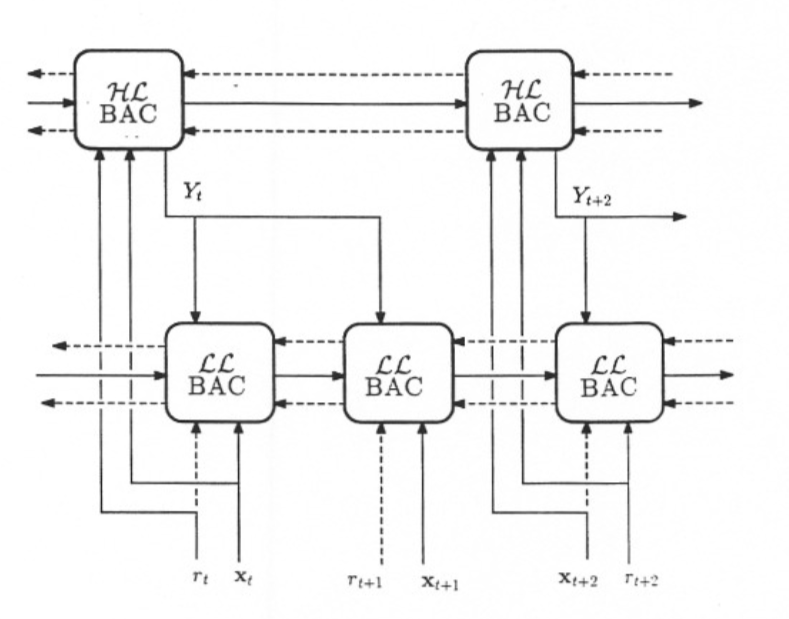}
\caption{\it Schematic of the Two-Level BAC}\label{fig-2LB}
\end{figure}

The \HL BAC is virtually identical to the ``standard'' BAC shown in Fig. 2,
except the \HL action, ${\bf Y}_t$,  is used as part of the input to the
\LL instead of determining the force on the cart. ${\bf Y}_t$ is referred to
as the ``plan''; in general it can be a vector, but for examples in this
paper it is just a scalar quantity. Note that the \HL receives the usual
reinforcement from the environment. 

Fig. 4 shows the architecture of the \LL BAC, which has the reinforcement
signal 

\myeq{ r'_t = (Y_{t-L} - \theta_t)^2 \label{eq-40}}

\noindent 
where $\theta_t$  is the angle of the pole from top center, and $L$ is the
number of elapsed time steps since the last \HL update.  In other words,
the \LL controller learns to perform actions (forces on the cart) that keep
the pole at an angle equal to the plan input. Note that the \HL action,
$Y$, is input to the \LL action {\em and critic}  networks.

\begin{figure}[htp]
\centering
\includegraphics[width=0.65\textwidth]{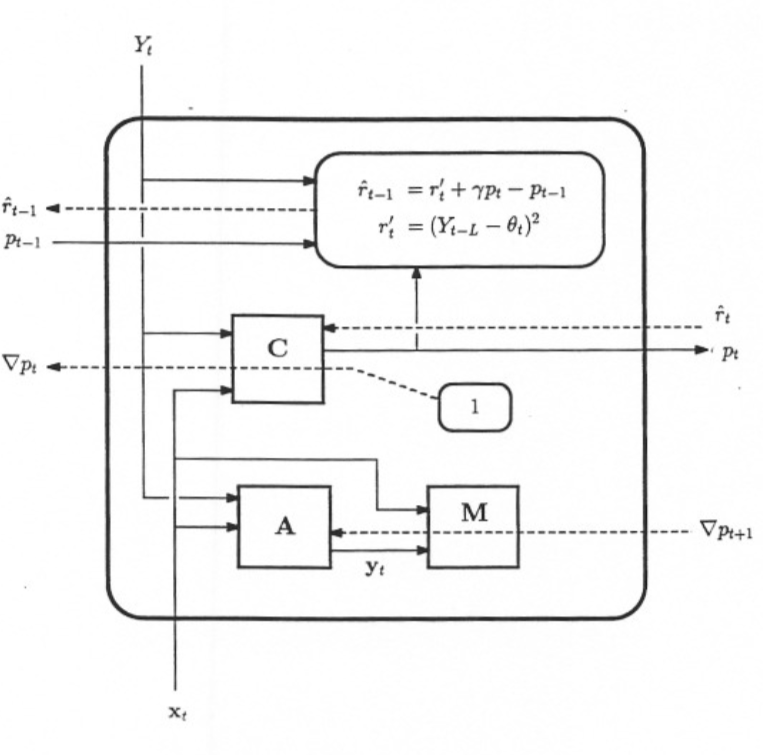}
\caption{\it Schematic of the Low-Level BAC (with explicit function)}\label{fig-LLB}
\end{figure}

It is most desirable that the environment of each BAC have the Markov
property, i.e., that future states and reinforcement signals depend only on
the current state and action(s). As Watkins pointed out
~\cite{watkins:1989}, if the \LL controller is subject to state transitions
that are not entirely caused by its own actions then the Markov property
does not hold for the \LL controller. However, the case here is a little
more subtle.  Suppose that, instead of predicting the value of the
cumulative reinforcement ($p$), the \LL critic predicted just $rÕ$ for the
next time step (this is equivalent to predicting p for $\gamma = 0$). 
Since the \LL critic has no way of knowing whether $Y$ is going to change
on the next time step, it would do a poor job of predicting $rÕ$ since the
latter depends on $Y$.  However, because $Y$ does stay constant for many
($N$)  \LL time steps between each change in its value,  the \LL criticÕs
prediction would be somewhere in the ``ball park;''  it would tend to yield
a short term average of the oncoming values of $rÕ$.  The fact that an
accumulation of $rÕ$ is predicted rather than just $rÕ$ significantly
mitigates the prediction difficulty because of the  averaging effect of the
accumulation.  

It was found for the simulations described below that the \LL controller
failed to work if the \HL action was not included in the \LL critic input
space. Otherwise the \LL learned to control the pole angle reliably.

Note that the \LL model network does not require the \HL action as part of
its input. This would not be the case, however, if the high level actions
affected the cart-pole system through some other channel than the
low level controller, such as through additional forces on the cart.

The single-level Indirect BAC (of the previous section) works best with two
phases of learning---the model network is trained in the first phase for
random actions, then the action network is trained via the feedback
from the critic. A similar situation occurs for the two-level BAC except
that two phases are utilized for each level. The phases are reiterated
below for clarity. 

\begin{description} 
\item[\ \ Phase I:]  The \LL model network learns to predict the change in
state for random \LL actions over time step. The \LL action weights are
then frozen.

\item[\ \ Phase II:]  The \LL action network learns to balance the pole
angle at an angle proportional to the \HL action (plan) for random values
of the latter. The \LL action weights are then frozen.
 
\item[\ \ Phase III:] The \HL model network is trained to predict the
change in state for random \HL actions N time steps ahead. The \HL model
weights are then frozen.

\item[\ \ Phase IV:]  The \HL action network learns to keep the cart
centered. 
\end{description}

For the two-level Direct BAC only {\em two\/} phases of learning are
required, where phases I and II for the Direct BAC correspond to phases II
and IV for the Indirect BAC, respectively. 

\subsection{Experimental Results}
Table 2 presents results obtained for the cart-pole system with a
servo rate of 50 hz---corresponding to the most difficult of the cases
presented in Table 1. Note that the performance of the \HL controller in
phase IV learning was virtually identical to the performance of the system
described in ~\cite{jameson:1990}, where a proportional derivative control
loop essentially fulfilled the role of the \LL controller in the present
example. The learning performance in phases I through III was very
reliable, i.e., no local minima were encountered over the ten experiments. 

\begin{table}
\center{
\begin{tabular}{||l|c|c|c||} \hline\hline \multicolumn{4}{|c|}{
\parbox{3in}{\center{{\bf TABLE 2}\\ Results for Two-Level BAC
Architectures\\ with Explicit Low-Level Function over Ten Experiments\\
(50 hz Servo Rate)
\medskip}}} \\ \hline  \hline \parbox{1.3in}{\ \\ \smallskip Architecture
\\ \smallskip} & Phase & \parbox{1in}{Typical No. of Trials} & 
\parbox{1in}{Average No. of Time Steps} \\ \hline
Indirect BAC & I & 800 & 10,000  \\ \hline 
"\ "\ " & II & 900 & 130,000  \\ \hline 
"\ "\ " & III & 400 & 150,000  \\ \hline 
"\ "\ " & IV & 160 & 10,000  \\ \hline \hline 
Direct BAC & I & 1,600 & 650,000  \\ \hline 
"\ "\ " & II & 450 & 50,000  \\ \hline
\multicolumn{4}{l}{} 
\end{tabular}
}  
\end{table}

One way to make a direct comparison between the results shown in
Table 2 for the two-level BAC with the corresponding single-level BAC   
(for the 50 hz servo rate) is to compare the total number of time
steps prior to success. This yields 300,000 steps for the two-level BAC
and 360,000 steps for the single-level BAC (10,000 steps are included in
the latter to account for phase I learning---which is not included in
Table 1). However, of the thirty experiments in Table 1, only {\em six\/}
succeeded, whereas the success ratio for the two-level BAC was much greater
(nine successes out of ten experiments for phase IV and no failures in the
prior phases). 

Given the additional complexity associated with the model network, a
reasonable conclusion might be that having a model is not worth the effort
for a two-level BAC. For the cart-pole problem this might be the case.
However, it seems likely that there are many problems where a Direct BAC
would be inadequate for a given single level of control. In fact, the
single-level BAC experiments of Table 1 indicate that the Indirect BAC was
far inferior to the Direct BAC for the 50 hz servo rate.

\section{Response Induction (RI) Learning \label{sec-50}}

What happens if, instead of forcing the pole to ``follow'' the \HL plan
through an internal \LL reinforcement, the \LL controller is provided the
same (external) reinforcement as the \HL?  What role does the \HL plan,
with its influence on the \LL controller, take on for this case, i.e.,
what ``subtask'' does the \LL controller learn?  I generally found that,
given enough time, the effects of the \HL plans on the \LL actions became
negligible. Since there was no inclusion of the plans in the reinforcement
provided to the \LL controller, the plans became disturbances to be
rejected. 

One way to {\em induce\/} the \LL to learn to react to the \HL plan signals
without expressing explicitly {\em how\/} to react to them is to affect the
weights of the action network such that each

\myeq{\delta_i\ \equiv\ {\partial p\over \partial Y_i},
\quad \forall\ i\in {\cal P}\label{eq-53}}

\noindent
is some significant positive value, where $p$ is the \LL critic output (see
eq. 1), $Y_i$ is the $ith$ \HL plan input, and ${\cal P}$ is the set of
units in the \LL action network corresponding to the plan inputs; note that
The term ``$\delta$'' is used in the same sense as Rumelhart etal.
~\cite{rumelhart:1986}, Vol. 1, Chapter 8, with respect to backpropagation.
Again, for this case the \LL BAC receives the same reinforcement signal
from the environment as the \HL BAC. 

Perhaps the more obvious approach for inducing a response would be to
replace  $\partial p / \partial Y_j$ with $\partial y_i / \partial Y_j$ in
eq. \ref{eq-53}, where $y_i$ and $Y_j$ correspond to the $ith$ and $jth$ \LL
and \HL action signals, respectively. Yet this was found experimentally to
be much less effective. The disturbing effect of the response induction
usually kept the \LL BAC from keeping the system stable for any appreciable
amount of time. This was probably because response induction based on the
action network output requires an immediate effect, whereas the  induction
based on the critic output elicits a response which can be mitigated over
several time steps (since the critic output reflects present and future
events). It is fortuitous that most of the factors for critic-based RI
learning are already computed during \LL action network learning. 

To see how the \LL action network weights are adapted to increase
$\delta_i$ (for $i\in {\cal P}$), consider the following error functon for
the output of the \LL action network:

\myeq{E\ =\ -E_{\hbox{\scriptsize\it critic}}\ + 
\ E_{\hbox{\scriptsize\it influence}},\label{eq-60}}  

\noindent  where 

\myeq{E_{\hbox{\scriptsize\it influence}}\ =\ -{k_1\over n_p}\ 
\sum_{i\in {\cal P}}  e^{-\delta^2_i/k^2_2},\label{eq-70}}

\noindent
where ${\cal P}$ is the set of input units corresponding to the plan
inputs, $E_{\hbox{\scriptsize\it influence}}$ is the ``influence error,'' 
and $E_{\hbox{\scriptsize\it critic}}$ is the effective error of the \LL
action output due to feedback from the critic network (during phase II
learning for the two-level Indirect BAC or phase I learning for the
two-level Direct BAC). The weights of the \LL action network are adapted
such as to maximize $E$ in eq. \ref{eq-60}. The constant $k_1$ determines
the maximal degree of the planÕs influence and the constant $k_2$
determines the rate at which the influence is induced (I 
generally obtained adequate results with $k_2 = k_1/2$; note that if $k_2$ is set
too large the maximal induction, $\delta_i\approx k_1$, will not be
achieved). 

For the experiments, the weight of each connection between units in the
hidden layer to the (single) plan input unit was adjusted an amount
proportional to $\partial E/\partial w_{ji}$ (the rest were adjusted
normally):

\myeq{\Delta w_{ji}\ =\ \eta\left[\delta_j Y_i\ +\
k_1\, k_2\, \delta_i \delta_j\, e^{-\delta^2_i/k^2_2}
\right]\ +\ \hbox{momentum},\label{eq-64}}

\noindent 
where $\eta$ is the learning rate, $w_{ji}$ is the weight connecting the
$jth$ hidden unit to the $ith$ input unit (the plan unit), and $\delta_j$
corresponds to the $jth$ hidden unit and is defined in the manner of
Rumelhart etal. 

With the approach just described the \LL action network essentially learns
to maximize its reinforcement while responding in some fashion to
the \HL plans (which changes every $N$ \LL time steps). Experiments with the
cart-pole system revealed that the \LL action network reliably produced
smooth and significant responses to the \HL plans while keeping the pole
balanced. At the very beginning of the \LL BAC learning $\delta_i$ was
typically small, and hence the rate of the response induction was small
because operation was in the central (flatter) region of the Gaussian
``bowl.'' This allowed the system to achieve reasonable stability before
stronger responses to the \HL actions were induced.

\begin{figure}[htp]
\centering
\includegraphics[width=0.85\textwidth]{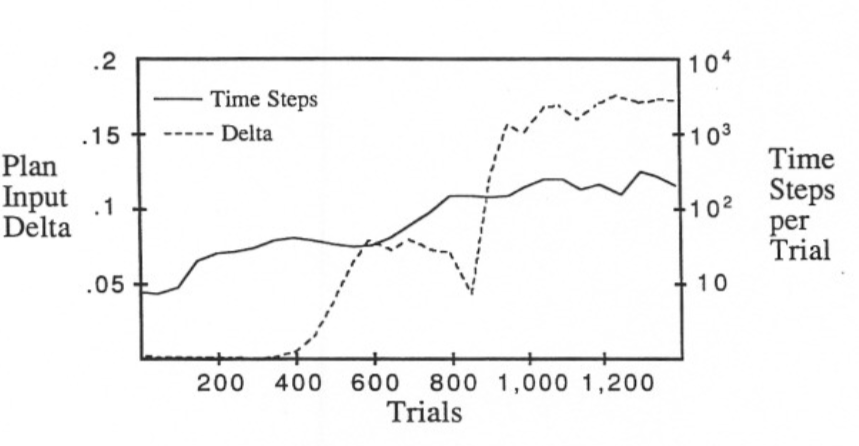}
\caption{\it Plot of ($\delta_i$) and the number of
steps/trial versus the number of trials}\label{fig-trials}
\end{figure}

Figure 5 shows the evolution of $\delta_i$  for three typical experiments
during phase II two-level Indirect BAC learning; note that $i=5$, i.e., the
first four inputs to the action network were the state variables, and the
fifth was the single plan input. Both curves were smoothed by averaging
the steps per trial (and the $\delta_i$) into bins of 50 trials. This plot
shows how the response induction occurs after moderate stability of the
system has been achieved.

I have ignored so far the implications of RI learning for case when the plan
input has more than one component. For this case, it is desirable that the
{\em response trajectory} due to each plan component be different than the
response trajectories due to the remaining plan components, i.e., that the
same response trajectory for one plan component value cannot be achieved for
some value of another plan component. The fact that the plan components vary
independently with respect to each other during RI learning does not
guarantee that the trajectories of the responses will be be different,
although it seems highly likely that they  will be somewhat different.
Finding methods which enhance the diversity of the response trajectories is
certainly one area for future research.

\subsection{Experimental Results with RI Learning}

During the experiments it was usually the case that a change in the plan
input caused a fairly rapid shift in the pole angle, after which the pole
moved smoothly from the new (shifted) position, usually in the direction
from which it started.  In several experiments, however, the \LL was
observed to balance the pole at an angle proportional to the plan signal,
just as in the more supervised case in Sec. \ref{sec-10}.

After performing some degree of (RI) parameter optimization, I
performed fifteen experiments using RI learning during phase II for the
two-level Indirect BAC; all other parameters were identical to those used
for obtaining the results given in Table 2. During phase II RI learning
learning and phase III \HL model learning, uniformly random \HL actions in
the ranges [-.3, .3] and [-.7, .7], respectively, were provided to the \LL
BAC. The values of $k_1$ and $k_2$ were .35 and .14, respectively.

Over these fifteen experiments, the average number of trials/(time steps)
deemed adequate for phase II RI learning learning was 1,300/190,000. The
average number of trials/(time steps) deemed adequate for phase III (\HL
model) learning was 1,000/140,000. For phase IV learning, five of the
experiments resulted in finding a solution within 700 trials (otherwise the
experiment was halted). The average number of trials/(time steps) for the
solutions was 470/25,000.   

A common problem observed during the (ten) failures was that the state
space was covered insufficiently during RI learning. This problem
was tacitly eliminated for the explicit role case of the previous
section in that the corresponding response did cover the solution space,
i.e., the random plan signals corresponded to random pole angles,
distributed around its top center position.

\subsection{Outlook on RI Learning}

Inducing the \LL action network simply to react to the plan signals while
maximizing its local reinforcement seems crude in the sense that the type
of reaction learned may not necessarily be useful for \HL  performance.
However, because it is so general it lends considerable leeway to the \LL
action network for reinforcement maximization. This leeway could be
utilized most effectively, perhaps, by augmenting the ``environmental''
reinforcement signal with terms related to such factors as state space
exploration and/or to trajectory smoothing. It may be useful to construct
separate critics solely for the  purpose of response induction, letting a
new critic respond to some of the factors just mentioned. In any case, for
most applications it seems most desirable that the \LL behavior with
respect to \HL actions be deterministic, and that the state space is
adequately covered.   

\section{Role Learning Based on Time Constant Identification}
Another potential way to induce useful \LL role emergence is more directly
related to the ``explicit function'' learning in Sec. 3. This would be to
have the intelligent controller figure out itself which components of the
plant's state variables are ``slow'' and which are ``fast'' and apportion
control appropriately (e.g., as the control was apportioned in Sec. 3).
This would not be as general as RI learning in that specific state variables
must be identified and routed to different levels of control, whereas in RI
learning different {\em functions\/} of the state variables may implicitly
be routed. However, it seems like this might be more practical than
RI learning for the near term. 

\section{Conclusion \label{sec-70}}
In this paper I have addressed the problem of extending the time span over
which causes and effects can be accounted for reliably by a Backpropagated
Adaptive Critic. The general ideas, of course, are applicable to other
incremental learning methods (e.g., the feedback in time method
~\cite{nguyen:1989}).

The functional allocation within the hierarchical architecture has been 
approached from two extremes: in single case, Explicit Role Learning (ERL),
the role of the low level controller is predefined based on knowledge of the
system, and in the other case, Response Induction Learning (RIL),  no
knowledge and very little structure is imbedded in the architecture. 
   
Results for the two-level BAC with ERL show that it is considerably more
reliable than a corresponding single-level BAC for the cart-pole problem,
although at a cost of more user interaction. It is likely that the model and
action/critic learning phases within each BAC level can be meshed together
in a form of learning that has attributes of both of the original
(different) phases.  This is certainly one area for future work (for 
single-level  Indirect BAC learning as well as the  hierarchical case). 
Elimination of the phases related to separate learning for the low and high
level BAC's seems to be fundamentally much more difficult because of the
desirability of surrounding each BAC (level) with a stationary environment.
Perhaps the most promise lies in providing mechanisms for automatically
detecting when to switch these phases, rather than finding ways to eliminate
the phases themselves.

RIL was introduced as an example of how the emergence of functional roles
might be achieved within a hierarchical (reinforcement) control system. The
method is very reliable at inducing responses while allowing the controller
to ``mind'' the reinforcement signals. However, such responses, while
usually quite deterministic, were not always conducive to solving the
problem. However, the generality of the method leaves room for many
varieties of enhancement.

The extension of the present approach to more than two levels of control
seems straightforward in principle. For this case, the servo rates of each
level would be progressively slower as the hierarchy is ascended, and each
level would control the level beneath it. Also, the present discussion has
also been limited to cases where control signals are issued at equal time
intervals. The extension to intermittent  control is another important area
for future research.  

Is this hierarchical BAC approach valid for more challenging problems
than the cart-pole system?  The difficulty with the latter seems not to be
with the control logic required (which is fairly simple), but rather with
the widely differing time constants of the motion of the cart and pole, a
hypothesis which is strongly corroborated by results in Table 1.  It is
likely that a single-level BAC can learn much more (logically) difficult
problems than the cart-pole as long as causes and effects occur within 
reasonable  time spans. The multi-level BAC is just a way of extending the
performance when large differences in time constants are present, whether
the ``logical'' problem addressed within each level is difficult or not.

\zapbib